\pdfoutput=1

\documentclass[11pt]{article}

\usepackage{emnlp2021}

\usepackage{times}
\usepackage{latexsym}

\usepackage[T1]{fontenc}

\usepackage[utf8]{inputenc}

\usepackage{microtype}

\usepackage{amsmath}
\usepackage{amssymb}
\usepackage{amsthm}
\usepackage{amsfonts}
\usepackage{array, booktabs, makecell}  
\usepackage{algorithm2e}

\usepackage{microtype}
\usepackage{graphicx}
\usepackage{subfigure}
\usepackage{url}
\usepackage{comment}
\usepackage{wrapfig}

\newcommand{\sD}{\mathcal{D}}
\newcommand{\sX}{\mathcal{X}}
\newcommand{\sY}{\mathcal{Y}}

\newcommand{\Ebb}{\mathbb{E}}

\title{MetaXT: Meta Cross-Task Transfer between Disparate Label Spaces\\
}

\author{Srinagesh Sharma \\
  Microsoft \\
  1 Microsoft Way \\
  Redmond, WA \\\And
  Guoqing Zheng \\
  Microsoft Research \\
  1 Microsoft Way \\
  Redmond, WA \\
  \texttt{\{srsharm, zheng, hassanam@microsoft.com\}} \\\And
  Ahmed H. Awadallah \\
  Microsoft Research \\
  1 Microsoft Way \\
  Redmond, WA 
  }

\begin{document}
\maketitle
\begin{abstract}
Albeit the universal representational power of pre-trained language
models, adapting them onto a specific NLP task still requires a
considerably large amount of labeled data. Effective task fine-tuning
meets challenges when only a few labeled examples are present for the
task. In this paper, we aim to the address of the problem of few shot
task learning by exploiting and transferring from a different task
which admits a related but disparate label space. Specifically, we
devise a label transfer network (LTN) to transform the labels from
source task to the target task of interest for training. Both the LTN
and the model for task prediction are learned via a bi-level
optimization framework, which we term as MetaXT. MetaXT offers a
principled solution to best adapt a pre-trained language model to the
target task by transferring knowledge from the source task. Empirical
evaluations on cross-task transfer settings for four NLP tasks, from
two different types of label space disparities, demonstrate the
effectiveness of MetaXT, especially when the labeled data in the
target task is limited.
\end{abstract}

\section{Introduction}
\label{sec:introduction}

Recent advances in large-scale self-supervised pre-training have
greatly impacted and shaped transfer learning for natural language processing. Namely, a two-stage process including pre-training with
self-supervision loss and fine-tune on task data~\cite{devlin2018bert,radford2019language,conneau2019unsupervised,yang2019xlnet,raffel2019exploring} has become a standard. Increasingly larger pre-trained language models provide
richer and more powerful representations for text reducing the need for large amounts of task labeled data~\cite{brown2020language}. However to attain the best task performance, adequate amount of task labeled data is still required. For example, the largest pre-trained language model to date, GPT-3~\cite{brown2020language}, achieves an average score of 70.3 on SuperGLUE~\cite{wang2019superglue} on few-shot setting, while the best supervised model, as the time of writing, scored 90.3, demonstrating the importance of task labeled examples even with large pre-trained models. 

Many NLP tasks may have limited amounts of training data, making it challenging to train performant models. However, it is often possible that a reasonably large dataset from a related task can be found - such related tasks are often referred to as auxiliary tasks. To alleviate the scarcity of task labeled data in low-resource tasks, many NLP methods leverage multi-task learning or cross-task transfer learning, e.g. ~\cite{ruder2017overview, balikas2017multitask, augenstein2018multi,pfeiffer2020mad}, to incorporate additional data from one or more auxiliary tasks into the training process.

\begin{figure}[t]
  \centering
  \includegraphics[height=0.2\textwidth]{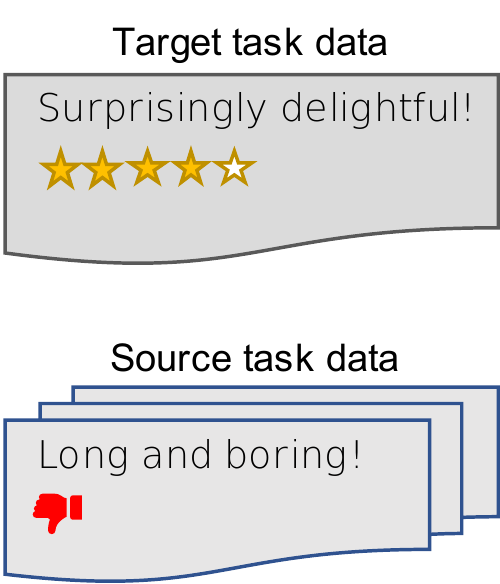}\hspace{0.4in}
  \includegraphics[height=0.2\textwidth]{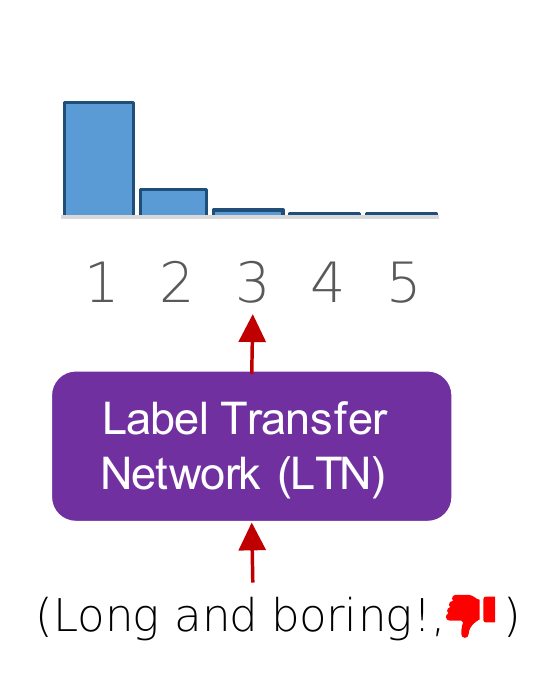}
  \caption{\textbf{(Left)} Example task pairs, where the target task,
    sentiment analysis on 5-star reviews, has limited labeled
    data and the source task, sentiment analysis on binary
    reviews, has enough labeled data (showing actual review titles);
    \textbf{(Right)} The label transfer network is proposed to learn
    the mapping from the binary sentiment onto a distribution over the 5-star review
    target task that the transferred labels can
    augment the target task data for a larger set of labeled
    examples.}
  \label{fig:demo}
  \end{figure}

In this paper, we focus on studying cross-task transfer from an auxiliary source task, where large amount of labeled examples might be readily available or easier to acquire, to a low-resource target task, where limited labeled data is available. Note that unlike traditional Multi-task learning~\cite{caruana1997multitask}, we do not attempt to learn multiple task jointly, rather we focus on improving performance on the low-resource target task using data from a  resource-rich source task. We focus on scenarios where the source task and the target task may be related, albeit having disparate label spaces (e.g. different label granularity, disjoint tag sets, etc.). Previous work has shown that similar tasks that have originated from similar probability distributions~\cite{caruana1997multitask} or have similar inductive biases ~\cite{baxter2000model} tend to perform well in multi-task learning settings. While no universal measure for task similarity exists, there has been several attempts at quantifying task similarity and its effect on transfer learning performance ~\cite{ruder2017overview,schroder-biemann-2020-estimating}.

Learning models for multiple loosely related tasks have mostly been studied in a multi-task joint training ~\cite{augenstein2018multi} or splitting parameters into shared and private spaces ~\cite{liu-etal-2017-adversarial}. We hypothesize that directly optimizing for the target task would result in better performance. However, this may be challenging given the disparate label space between the source auxiliary task and the target task. To overcome this challenge, we propose a meta-learning based method, MetaXT, to bridge the gap between the disparate label spaces and allow for effective transfer to low-resource tasks. MetaXT {\em learns to transfer} the label space from the source task to the target task in a way that maximally facilitates transfer to the low-resource target task using a label transfer network (LTN). Directly training the LTN in a supervised fashion is not possible due to lack of parallel data between the source and target tasks; instead we adopt a data-driven approach to learn both the main model architecture and the LTN via a bi-level optimization formulation. Figure \ref{fig:demo} shows an illustration for an actual task setting and the purpose of LTN.

MetaXT enjoys the following advantages:
\begin{itemize}
  \item The LTN learns a direct mapping from source task labels to
    target task, allowing explicit transfer across types, even when
    they come from disparate label spaces;
  \item The data-driven approach of learning offers flexibility in the transfer between tasks, which makes the method generally
    applicable to a wide range of NLP task pairs.
\end{itemize}
We conduct empirical evaluations of MetaXT over different data sets
and task pairs; comparisons with several multi-task training based
methods verify the effectiveness of the proposed approach. 


\section{Background and Preliminaries}
\label{sec:background}

Given a few-shot transfer learning setting, the target task has a
$k$-shot training set \(D_t := \{(x_1, y_1), ..., (x_{k}, y_{k})\}\),
\(x \in \sX_t, y \in \sY_t\), which is also augmented with labeled
examples from a source task \(D_s := \{(x_1, y_1), ..., (x_{N},
y_{N})\}\), \(x \in \sX_s, y \in \sY_s\) such that \(k \ll N\). Note
that in general $\mathcal{Y}_s$ is different from $\mathcal{Y}_t$,
meaning that the target task label space is disparate from that of the
auxiliary task. The auxiliary task is also referred to as the source
task, as we look to improve the target task by transferring knowledge
from it. Existing works on multi-task learning often
involves building on top of a text feature encoder $f_\theta$, often a
pre-trained language model, and adding task-specific layers to map
the text representations onto the respective label spaces,
parameterized by $v$ and $w$ for the source and target tasks,
respectively. For simplicity, we denote the complete
source task model and target task model as $f_{\theta, v}$ and
$f_{\theta, w}$, respectively, where $\theta$ denotes the weights of
the pre-trained language model shared by both tasks.

The classic multi-task joint training then aims to minimize the
combined loss over a set of tasks, hoping that the model could benefit
from other tasks. A recent line of work on
adaptors~\cite{pfeiffer2020adapterfusion,pfeiffer2020mad} builds on
top of multi-task joint training by training and leveraging task or
language specific expert models, however these still lie in the
category of multi-task joint training. One major difference between
multi-task joint learning and cross-task transfer for a particular
target task is that, multi-task joint learning aims to improve over
all tasks while cross-task transfer only aims to improve the target
task of interest where the other tasks are only used as auxiliary
information to improve the target task performance. Recent work on
measuring the similarities among NLP tasks and identifying the correct
task to use as auxiliary task for a particular target task reveals the
advantage of leveraging auxiliary tasks for transfer learning\cite{schroder-biemann-2020-estimating}.


\section{MetaXT: Meta Cross-task Transfer}
\label{sec:method}

\begin{figure*}[t]
  \centering
\includegraphics[height=0.25\linewidth]{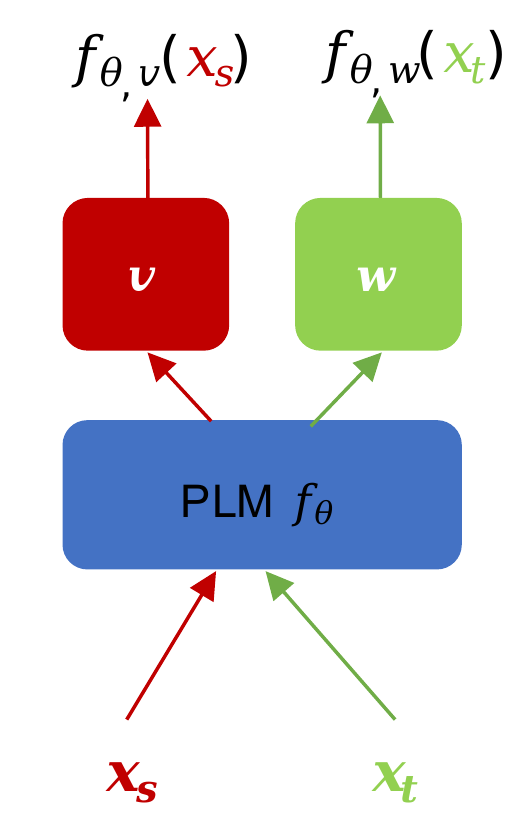}\hfill
\includegraphics[height=0.25\linewidth]{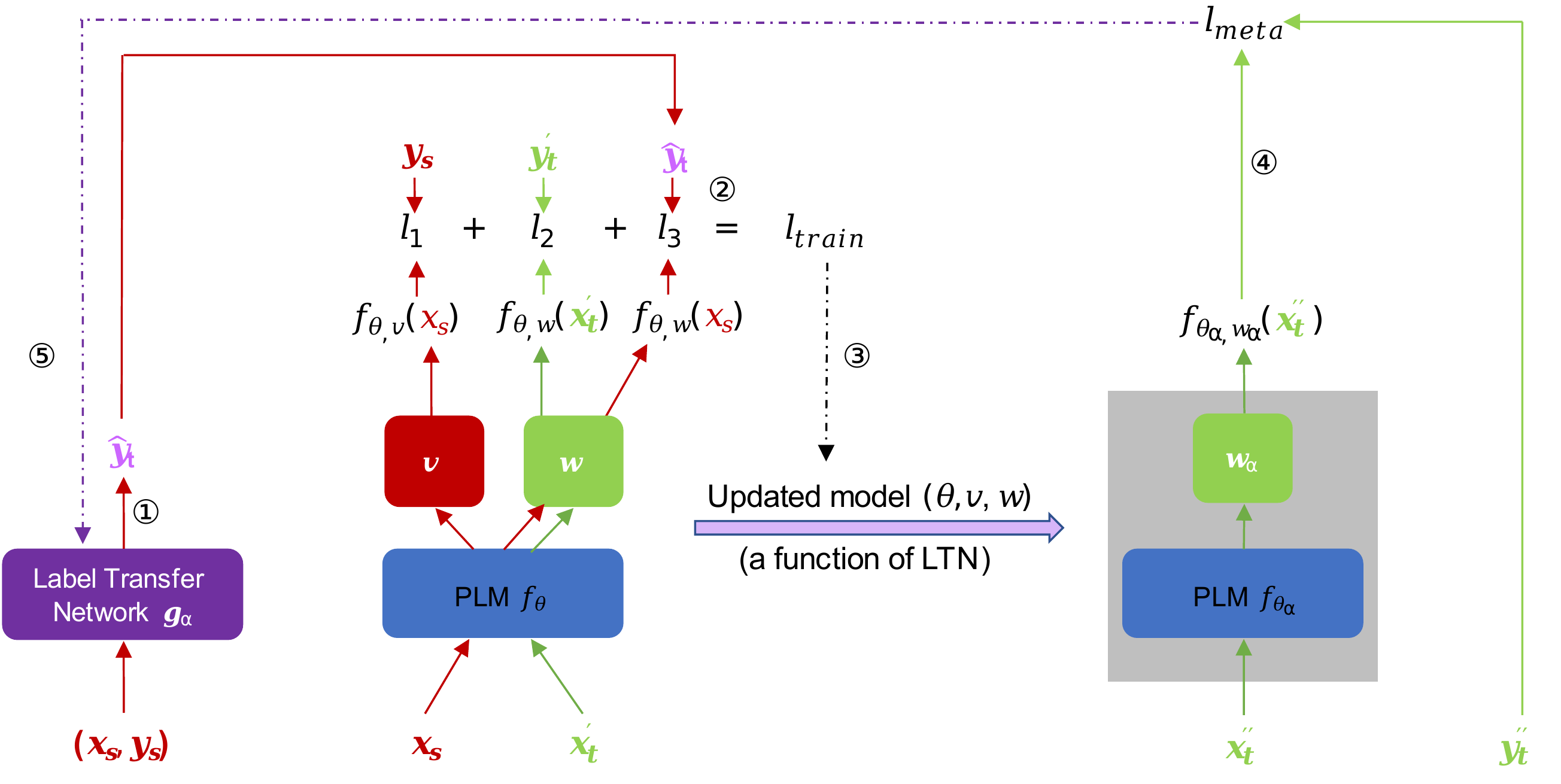}\hfill
\includegraphics[height=0.25\linewidth]{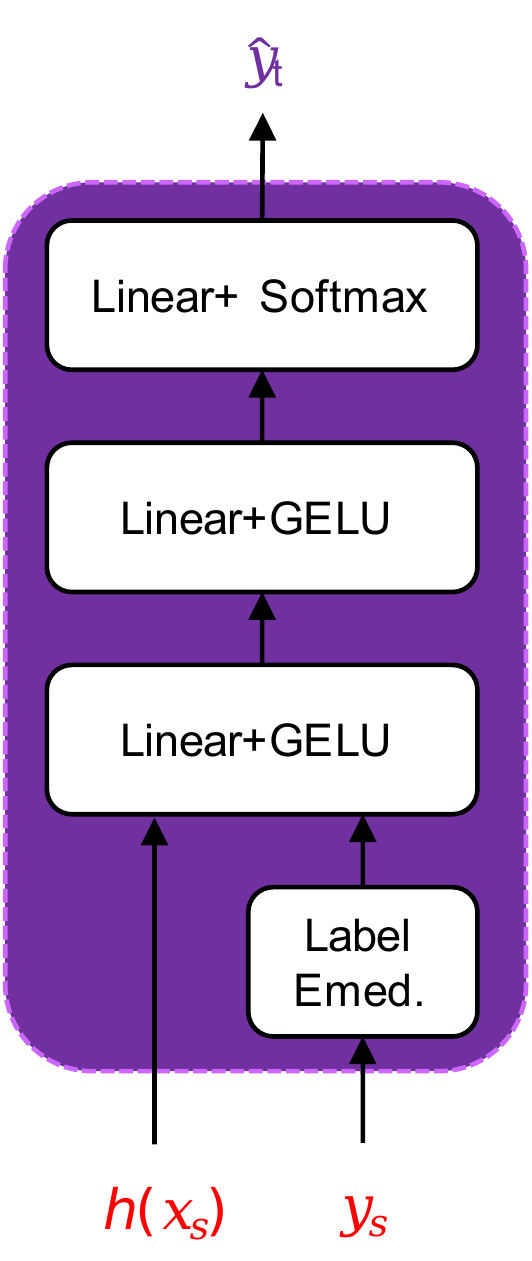}  
\caption{\textbf{(Left)} \textbf{Multi-task training}. Data examples from two
  tasks ($x_s$ for source and $x_t$ for target) are fed to the same
  representation network, i.e., a PLM $f_\theta$, and two task
  predictors are placed on top, parameterized by $v$ and $w$ for the
  source and target task, respectively. All parameters $(\theta, v,w)$
  are trained by minimizing the joint loss over source task labels
  $y_s$ and target ones $y_t$; \textbf{(Middle)} \textbf{Training flow of
    MetaXT}. MetaXT has the same model design with multi-task
  training, except for an additional lightweight Label Transfer Network
  (LTN) $g_\alpha$ to explicitly transfer source task labels to target
  task. \textbf{Note that $(x_t', y_t')$ and $(x_t'',
    y_t'')$ are two sub-splits from the same training batch $(x_t,
    y_t)$ as used by multi-task training.} (Refer to the text for
  detailed description for each step); \textbf{(Right)} LTN architecture.}
\label{fig:model}
\vspace{-0.1in}
\end{figure*}

In traditional multi-task learning settings, all tasks
are jointly learned via minimizing the combined loss over them.
However, this might be sub-optimal as severe data imbalance between the
two tasks might render the learned model to either favor the source
task for its larger training size or over-fit on the limited labels from the 
target task. Instead in this paper, we propose to leverage the source
task explicitly by transferring from its label space onto the target
task, in a way such that the produced pseudo label can directly benefit
the target task. Specifically, we devise a label transfer network
(LTN) $g_\alpha:\mathcal{X}_s, \mathcal{Y}_s\rightarrow
\mathcal{Y}_t$, a component parameterized by $\alpha$ which takes a
source example as input and produced a transferred label in the target
label space. Specifically, for a labeled example in the source task,
the LTN attempts to estimate its corresponding \textbf{soft} label in
the target task space by $\hat y_t=g_\alpha(x_s, y_s)$.

Directly training such LTN in a supervised manner is typically
infeasible as in general the parallel correspondence between labeled
examples from the source task and target task is unknown. Instead, we
follow a simple intuition to train the LTN that \textit{an effective
  LTN should produce accurate transferred labels in target task label
  space from examples in the source task, which further allows us to train
  a good target task model such that it incurs low evaluation loss on
  a separate set of target task examples}. This can be formulated as
the following bi-level optimization problem:
\begin{align}
\label{eq:opt}  
  &\min_\alpha  \mathcal{L}_{meta}(\theta_\alpha^*,w_\alpha^*)\\
  \mbox{s.t. } & \theta_\alpha^*, w_\alpha^*, v_\alpha^*=\arg\min_{\theta,w,v}\mathcal{L}_{train}(\theta, w, v, \alpha)\nonumber
\end{align}
where
\begin{align}
  \label{eq:loss_meta}  
  \mathcal{L}_{meta}\triangleq& \Ebb_{(x, y) \in \sD_{t}} \ell(f_{\theta_\alpha^\ast,w_\alpha^\ast}(x), y) \\
  \label{eq:loss_train}
  \mathcal{L}_{train}\triangleq& \Ebb_{(x, y) \in \sD_{t}} \ell(f_{\theta,w}(x), y) \nonumber\\
   +& \gamma_1 \Ebb_{(x, y) \in \sD_s} \ell(f_{\theta,v}(x), y) \nonumber\\ 
   +& \gamma_2 \Ebb_{(x, y) \in \sD_s} \ell(f_{\theta,w}(x), g_{\alpha}(x,y))
\end{align}
The upper component of the above problem aims to minimize the evaluation
loss $\mathcal{L}_{meta}$ of the trained target task model on the
few-shot test examples for the target task, while the lower component
specifies that the target task model $f_\theta,w,v$ is trained by
minimizing a summed loss of three terms, listed as follows
respectively
\begin{itemize}
\item $\ell(f_{\theta,w}(x_t), y_t)$ encodes the supervised loss over the target task;
\item $\ell(f_{\theta,v}(x_s), y_s)$ encodes the supervised loss over the source task;
\item $\ell(f_{\theta,w}(x_s), \tilde y_t)$ encodes the supervised
  loss over the transferred labels produced by the LTN, where $\hat 
  y_t=g_\alpha(x_s, y_s)$.
\end{itemize}
Hyper-parameters $\gamma_1$ and $\gamma_2$ controls the balances among
the three terms, and they are to be tuned by cross-validation.

Note that it is crucial for $\hat y_t$ to be a soft label to allow
gradient propagation back from $\mathcal{L}_{meta}$ all the way back
to the LTN parameters, which can be instantiated by using a softmax as
the output layer for the LTN (We defer the detailed design for LTN
used in this paper to Section \ref{sec:ltn}). For sentence
classification and sequence tagging problems, cross-entropy (CE) loss is used for
$\ell$ which can be easily extended to support soft labels 
as $-\sum_{i}y_i\log p_i$
where $y$ is a soft-label with $\sum_i{y_i}=1$ and each $y_i\geq 0$,
and $p$ is a probability vector for a model's output. Note that it
boils down to classic cross-entropy loss for a hard label $y$ if $y$
is expressed in its one-hot form.

\subsection{Label Transfer Network (LTN)}
\label{sec:ltn}
Recall that the LTN is a function taking a source example pair and
producing a pseudo-label in the target label space. To ease the design
complexity of LTN and to add minimal parameter overhead from the task
model, we devise the LTN as 3-layer feed-forward network, whose input
 dimension is $h_{dim}+z_{dim}$, where $h_{dim}$ is the
representation dimension of a data example and $z_{dim}$ is the
embedding dimension to encode the discrete source task label. The LTN
outputs a vector with dimension equal to the number of classes in the
target task. With a pre-trained LM (PLM), like BERT, serving as the
text representation encoder as typically used various NLP tasks. In
practice, we take the contextualized representation from the
\texttt{[CLS]} token for a source task example $x_s$ to feed into the
LTN for sequence-level tasks (or the token representation for
token-level tasks). This also relieve the LTN from the burden of
encoding the raw text thus making the LTN lightweight. Refer to Figure
\ref{fig:model}(c) for an illustration of LTN used throughout this paper.

\subsection{Model Learning for MetaXT}

Similar to many bi-level optimization problems, analytically solving
Eq. (\ref{eq:opt}) is infeasible, as every change on the LTN
parameters $\alpha$ requires solving for the optimal solution to lower
optimization problem. Instead, a widely used strategy to approximate
the solution to the lower problem with a one-step SGD estimate~\cite{finn2017model,liu2018darts, shu2019meta, zheng2021meta}. To be
concrete, we solve the following proxy problem
\begin{align}
  &\min_\alpha \mathcal{L}_{meta}( \Theta(\alpha))\nonumber\\
  \mbox{s.t. }  &  \Theta(\alpha) = \Theta - \eta\nabla_\Theta \mathcal{L}_{train}(\Theta,\alpha)
  \label{eq:opt2}
\end{align}
where $\Theta$ is a shorthand for the group of main parameters
$\{\theta, w, v\}$ and $\eta$ is the learning rates for them. Computing the meta-gradient $\nabla_\alpha
\mathcal{L}_{meta}(\Theta(\alpha))$ is the essence of training MetaXT,
which can be obtained as follows
\begin{align}
  &\nabla_\alpha\mathcal{L}_{meta}(\Theta(\alpha)) \nonumber\\
  =&\nabla_{\alpha}\mathcal{L}_{meta}( \Theta-\eta\nabla_{ \Theta}\mathcal{L}_{train}( \Theta,\alpha))\nonumber\\
  =&-\eta \nabla_{ \alpha,   \Theta}^2\mathcal{L}_{train}( \Theta,\alpha)\nabla_{ \Theta'}\mathcal{L}_{meta}( \Theta')\nonumber\\
  =&-\eta \nabla_{ \alpha}\Big(\nabla_{  \Theta}^\top\mathcal{L}_{train}(\Theta,\alpha)\nabla_{ \Theta'}\mathcal{L}_{meta}( \Theta')\Big)
  \label{eq:sgd}
\end{align}
where 
$\Theta'= \Theta-\eta\nabla_{ \Theta}\mathcal{L}_{train}( \alpha, \Theta)$.
Detailed procedure to train MetaXT is in Algorithm \ref{alg}.

\begin{algorithm}[t]
  \caption{MetaXT Training procedure}
  \While{not finished}{
    Sample source task batch $(x_s, y_s)$ and target task batch $(x_t, y_t)$ \\
    Split target task batch $(x_t, y_t)$ into $(x_t', y_t')$ and $(x_t'', y_t'')$ \\
   Update LTN parameters $\alpha$ by descending Eq. (\ref{eq:sgd}) \\
   Update model parameters $\Theta$ by descending $\nabla_{\Theta}\mathcal{L}_{train}(\Theta, \alpha)$
 }
 \label{alg}
\end{algorithm}

Eq. (\ref{eq:sgd}) shows how to compute the gradient of the LTN parameters $\alpha$ (or meta-parameters). Following Figure
\ref{fig:model}, for each round updating the meta-parameters $\alpha$,
the steps are as follows:
\begin{itemize}
  \item[\textcircled{1}] Data example pair from
source task $(x_s, y_s)$ is fed to the LTN and an estimate of the
corresponding target label $\hat y_t=g_\alpha(x_s,y_s)$ is produced;
\item [\textcircled{2}] Now with three data and label pairs, i.e, $\{(x_s,
y_s), (x'_t, y'_t), (x_s, \hat y_t)\}$, the joint training loss
$\mathcal{L}_{train}$ is computed following Eq. (\ref{eq:loss_train});
\item [\textcircled{3}] Back-propagate $\mathcal{L}_{train}$ onto and update
the main model parameters $(\theta, v, w)$, which are functions of the
LTN parameters;
\item [\textcircled{4}] With the updated model, compute the
meta-evaluation loss $\mathcal{L}_{meta}$ on a separate set of target
task examples $(x_t'', y_t'')$ following Eq. (\ref{eq:loss_meta}),
which at its core is also function of LTN parameters $\alpha$;
\item [\textcircled{5}] Back-propagate $\mathcal{L}_{meta}$ onto the LTN
parameters $\alpha$ and update the LTN.
\end{itemize}
The updates of the main parameters
$(\theta, v,w)$ are the same as standard multi-task training. At inference time, the LTN is not used and the trained main model with the target task predictor is used.

\section{Experiments}
\label{sec:exp}

We conduct experiments to evaluate MetaXT 
over a range of data sets consisting different NLP tasks, comparing with a set of relevant baselines (Code for MetaXT will be made publicly available).
\subsection{Datasets}
Our goal is to study cross-task transfer learning between disparate label spaces, therefore we select data sets according to the following two criterion:
\begin{itemize}
    \item \textbf{Tasks with different label granularities}. This involves tasks of the same type but with different label granularities. For example, sentiment analysis with polarity (positive or negative) and that with multi-scale scores constitute one such pair of task. We pick two review data sets for this category, Yelp and Amazon. Note that for this setting we only evaluate transferring from polarity sentiment analysis to multi-scale sentiment analysis, as the reverse problem can be naively solved by thresh-holding the multi-scale scores to get the optimal polarity sentiment.
    \item \textbf{Tasks with different label types}. A more challenging setting than transferring between different label granularities is to learn to transfer from one type of task to a different type, for example, from part-of-speech (POS) tagging to Multi-Word Expressions(MWE) and between two related NER tag sets. For the first example, we transfer between POS tagging in UD English dataset to MWEs available in Streusle \cite{schneider2015corpus} and for the second one we transfer from the Conll2003 (German) corpus to the GermEval2014 \cite{benikova2014nosta} NER task.
\end{itemize}
Table \ref{tab:datasets} presents the statistics of all data sets used for evaluation. It is worth noting that, for a target task of interest  it is possible to find better auxiliary task to transfer from as pointed out by ~\cite{schroder-biemann-2020-estimating}. However this is orthogonal to this paper, where we aim to address better cross-task transfer from a methodological perspective, and thus finding the best source task is beyond the scope of this paper and thus left as future work.

\begin{table}[t]
\small
  \caption{Data set and task overview. (PS: Polarity Sentiment, MS:
    Multi-scale Sentiment, POS: Part-of-Speech, MWE: Multi-Word Expr., NER:
    Named Entity Recognition)}
  \begin{tabular}{cccc}
    \toprule
    \multicolumn{2}{c}{Source} & \multicolumn{2}{c}{Target}\\
    Dataset & Task(\#labels) & Dataset & Task(\#labels) \\
    \midrule
    Yelp Pol &  PS(2) & Yelp Full & MS(5) \\
    Amazon Pol & PS(2) & Amazon Full & MS(5) \\
    UD English & POS(17) & Streusle & MWE(3) \\
    Conll03(Germ) & NER(8) & GermEval14 & NER(25)\\
    \bottomrule
  \end{tabular}
  \label{tab:datasets}
  \end{table}

\subsection{Baselines}

We aim to compare MetaXT with a comprehensive set of baseline methods that are relevant to the settings. Since the problem we study involves multiple tasks, we consider the following set of applicable baseline methods (for details of the experimental setup please refer to Appendix A):
\begin{itemize}
    \item \textsf{Target Only}, which trains only on the few-shot target task examples. Note that a source only baseline is not applicable due to the different label spaces between source and target tasks;
    \item \textsf{Multi-task}, which jointly optimizes to learn both the source and target task with two task prediction heads put on top of the textual feature encoders;
    \item \textsf{AdapterFusion}, a recent work on building multi-task learners based on separately trained adapter modules for each task and then a fusion layer is trained to aggregate knowledge from all the task adapters~\cite{pfeiffer2020adapterfusion}.
\end{itemize}
It's worth noting that even though MAML~\cite{finn2017model} also addresses transfer learning between tasks with meta learning, it is not readily applicable for this setting as it aims to transfer and generalize from a set of source tasks onto new tasks \textbf{unseen} in training, i.e., the target task is unknown to the model in training, while in this paper we always assume that the target task is available for training. 
Additionally, MAML explicitly requires that all tasks at hand have the same dimensionality in their label spaces (even though they are different tasks). 

\subsection{Main Results}

\subsubsection{Task transfer with different label granularities}

\begin{figure}[t]
\centering
  \includegraphics[width=0.9\linewidth]{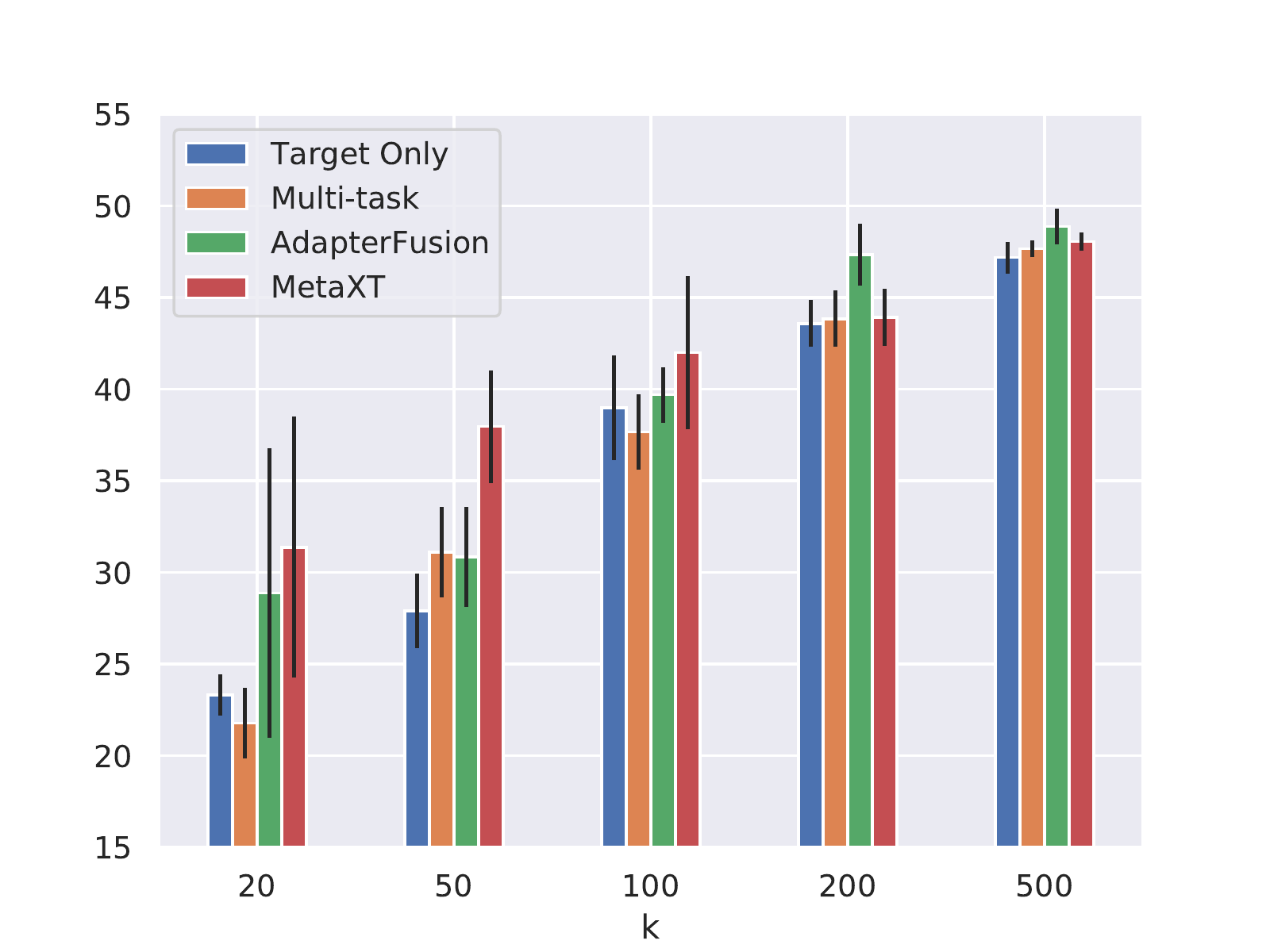}
  \caption{Accuracies on Yelp data. (Avg. from 5 runs)}
  \label{fig:yelp_main}
  \vspace{-0.2in}
\end{figure}

\textbf{Yelp Dataset:} The performance of the various models on the
Yelp Datasets are shown in Figure \ref{fig:yelp_main} for 5 different
number of labeled examples $k$ in the target task, i.e.,  $k\in\{20, 50, 100, 200,
 500\}$. 

Overall, while all methods benefit from a larger amount of target task data, they behave differently for
a given $k$. The \textsf{Target Only} baseline performs only marginally
better than random guess when $k$ is small $(k=20)$. The
\textsf{Multi-task} and \textsf{AdapterFusion} baselines perform
better by leveraging similarities with the source task. 
However, due to the limited number of labeled examples from the
target task, multi-task joint training is unable to effectively
counter the imbalance between source and target tasks, as shown by
$k=20$ and $k=100$. 
Instead, MetaXT is able to effectively
leverage the estimated target labels generated by LTN from source
task examples as additional supervision for training a target task
model, particularly when the number of target task examples $k$ is
small. Interestingly, the smaller $k$ is, the larger gain MetaXT
achieves over the \textsf{Target Only} and \textsf{Multi-task}
baselines, demonstrating the effectiveness of MetaXT for few-shot
transfer learning settings. As more number of target task labeled examples
are available, e.g. $k=200$ or $500$, the gaps between all the
methods shrink. This is expected, as transferring from the source task
might no longer be required.



\begin{figure}[t]
  \centering
  \includegraphics[width=0.9\linewidth]{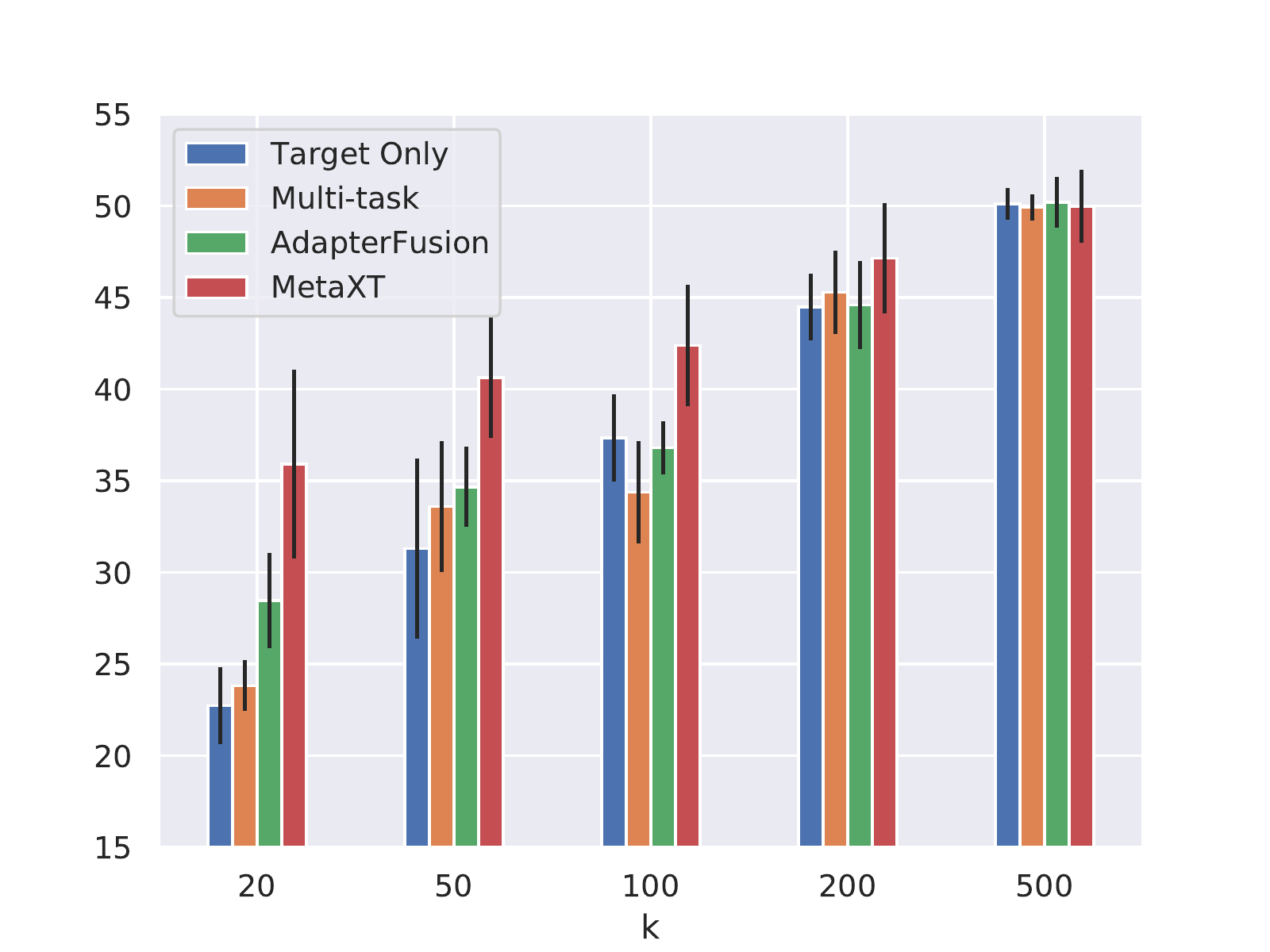}
  \caption{Accuracies on Amazon data. (Avg. from 5 runs)}
  \label{fig:amazon_main}
  \vspace{-0.2in}
\end{figure}

\textbf{Amazon Dataset:} The performance of MetaXT on product reviews
in the Amazon dataset is similar to that on the Yelp dataset , as
shown in Figure \ref{fig:amazon_main}. 
In addition, we
observe that MetaXT achieves significantly higher gains over all the
baseline methods when $k$ is small, i.e, $k\leq 200$, which
again validates that MetaXT is well suited for the few-shot learning
setting for task transfer. 

\subsubsection{Task transfer with different label types}

\begin{figure}[t]
\centering
  \includegraphics[width=0.9\linewidth]{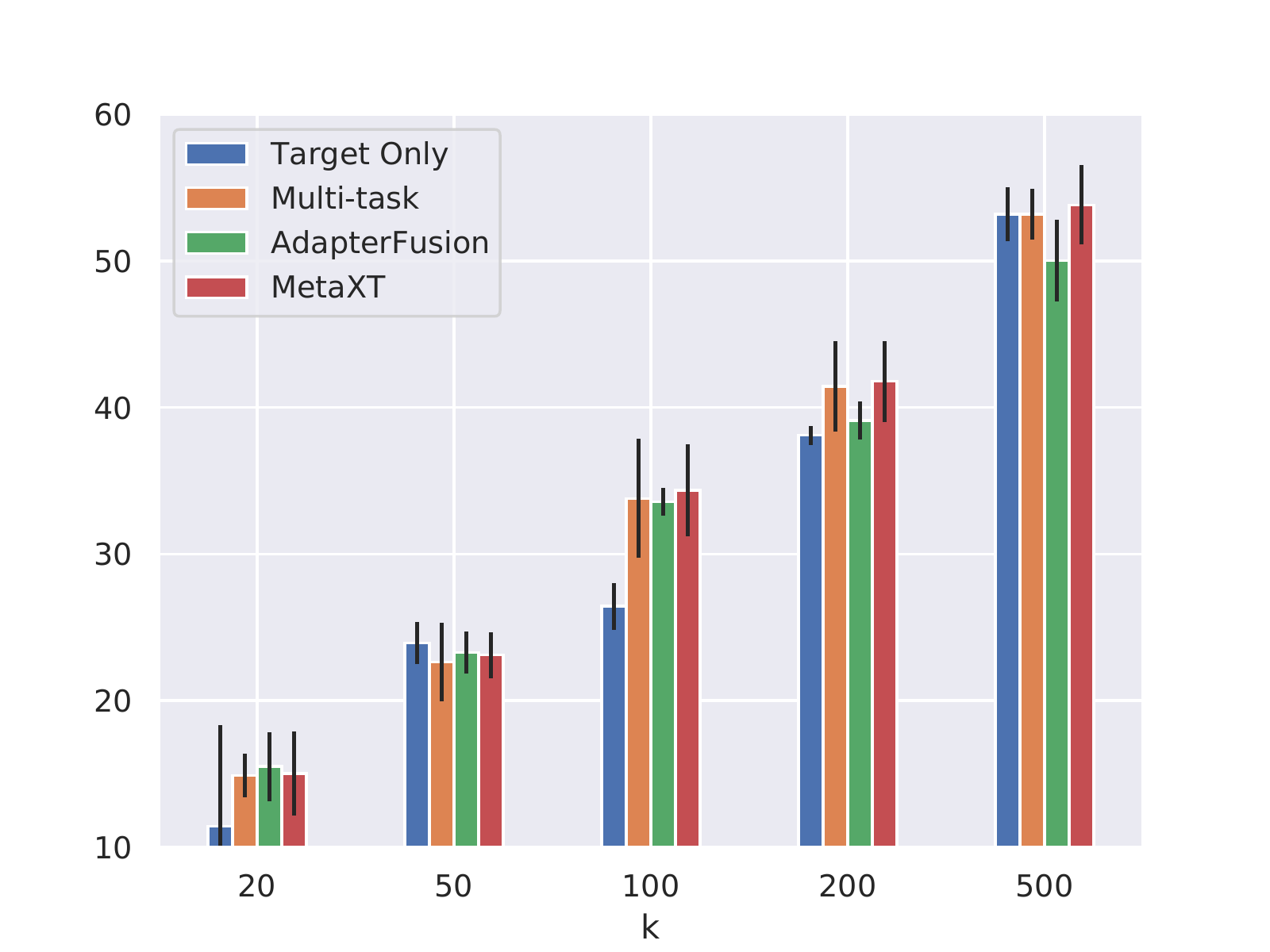}
  \caption{F1 Scores - POS to MWE. (Avg. from 5 runs)}
  \label{fig:streusle_main}
  \vspace{-0.2in}
\end{figure}

\begin{figure}[t]
\centering
  \includegraphics[width=0.9\linewidth]{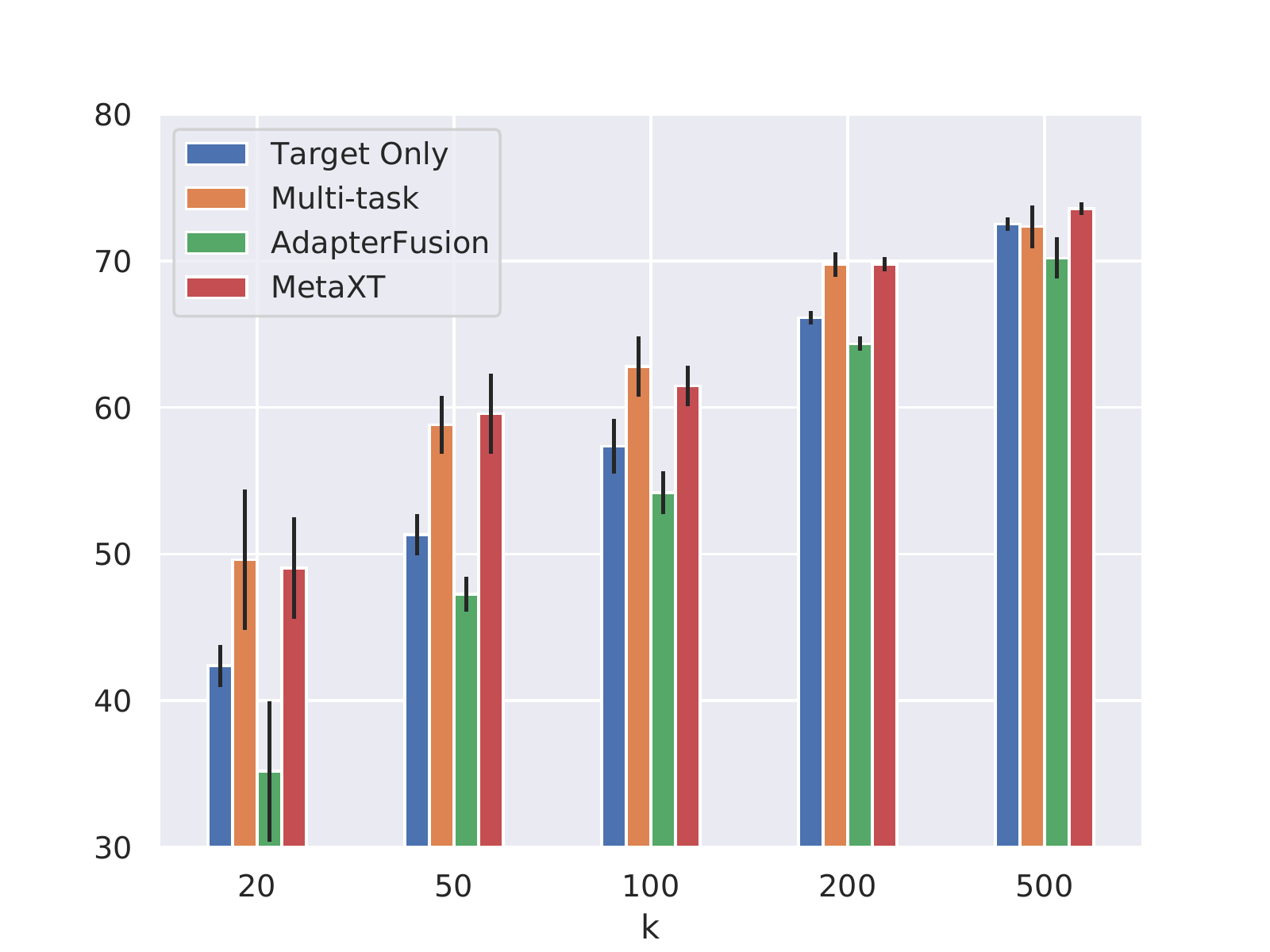}
  \caption{F1 Scores of NER Task. (Avg. from 5 runs)}
  \label{fig:ner_main}
  \vspace{-0.2in}
\end{figure}

\textbf{POS Tagging to MWE} We demonstrate transfer from
Parts-of-Speech (POS) tagging from the Universal Dependencies (UD)
English Tree Bank to MultiWord Expression (MWE) identification in
Streusle (Streusle 4.0 Dataset, \cite{schneider2015corpus}. We use the "standard" POS tagging with 17 classes provided
by the UD Dataset. For MWE, we use simplified BIO tagging 
of the Strong MWEs in Streusle,
similar to \cite{changpinyo2018multi}. The results
are as shown in Figure \ref{fig:streusle_main}. 

In this example, to adapt to the different domains of the source and target, we used a representation transfer network (RTN) \cite{xia2021metaxl} at the sixth layer of the transformer encoder in addition to the label transfer network. The representation transfer network was a 3 layer feed-forward network used only to adapt the source representation to the target while optimizing for the label transfer. This results in a modification of the 3rd term in Equation \eqref{eq:loss_train} to \(\gamma_2 \Ebb_{(x, y) \in \sD_s} \ell_w(f_{\theta,w, \phi}(x), g_{\alpha}(x,y))\), where \(\phi\) represents the parameters of the RTN.


For this particular transfer setting on token level classification
tasks, we observe that joint-training based methods, including
Multi-task and AdapterFusion, helps the target task by leveraging the
source task. MetaXT is able to achieve the best results on 3 out of 5
settings, while being close to the best performing methods for other
cases. Additionally, MetaXT is able to handle the additional challenges
brought by token-level classification tasks, as well as the data domain
shift, i.e. from English Tree Bank to Streusle.

\textbf{Transfer between related NER tag sets} We also demonstrate
transfer between two different but related NER tag sets - Conll 2003
German and GermEval2014\cite{benikova2014nosta}. GermEval2014 NER tag set is a superset of the
NER tagsets provided in Conll2003 by adding derived, partitioned and
other labels. Similar to the MWE task, we use an RTN at the sixth layer of the encoder to adapt the source domain to the target domain.
The F1 scores for the transfer task are as shown in
Figure \ref{fig:ner_main}. It can be seen
that the standard multitask model and the MetaXT models perform better
than the Target Only model, where as the AdapterFusion models do not perform
competitively in comparison to full fine-tuning for transfer between
tag sets. 

\subsection{Analysis and Ablation Studies}

\subsubsection{What exactly does the LTN learn?}

\begin{figure}[t]
    \centering
    \includegraphics[width=0.3\textwidth]{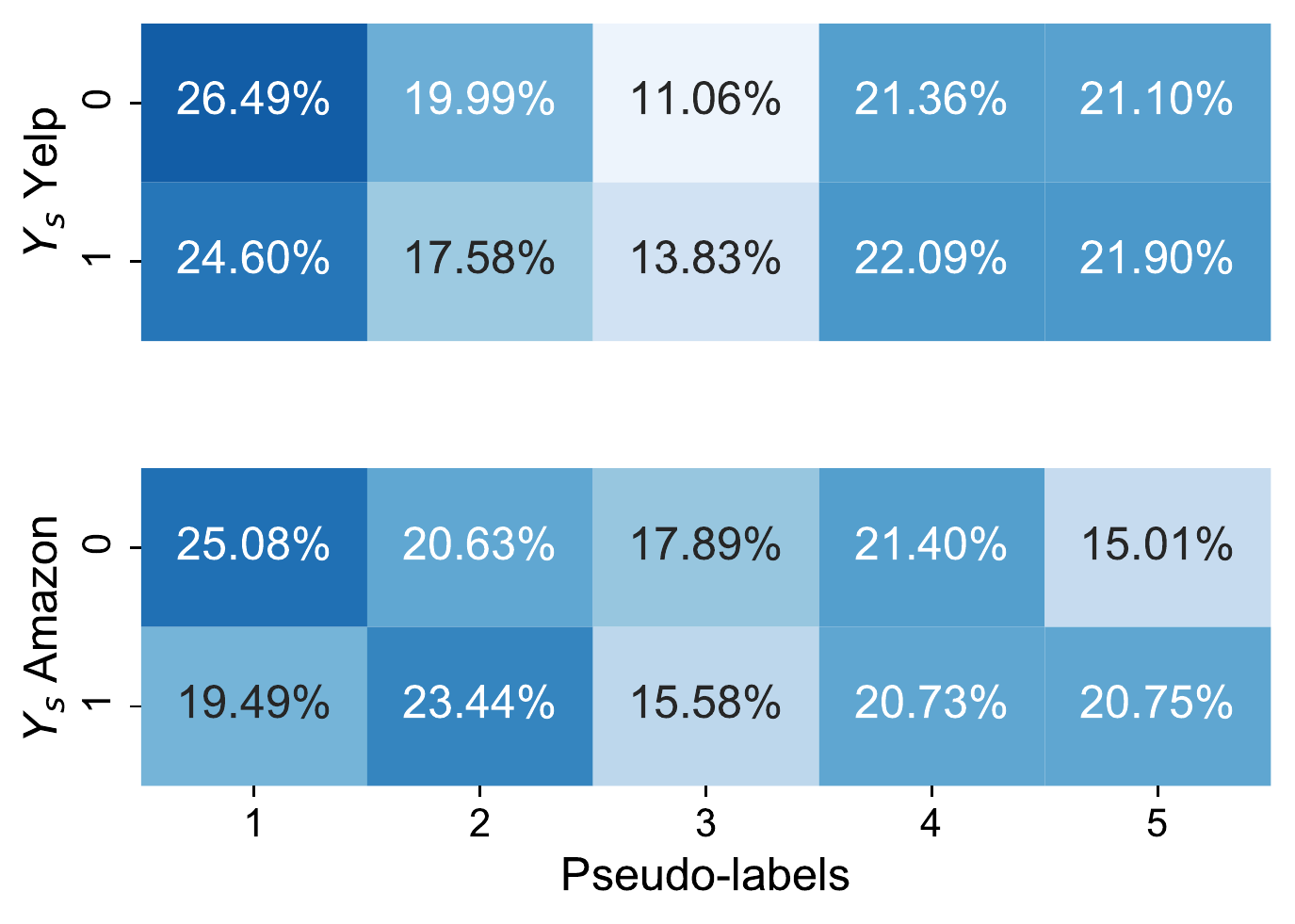}
    \caption{Distribution of \(\hat{Y}_t\) with \(Y_s\) for yelp and amazon datasets with 100 target examples. The source labels do not contain information on target label \(\{3\}\) and LTN is able to de-emphasize this.} 
    \label{fig:yelp_amazon_confusion_matrix}
    \vspace{-0.2in}
\end{figure}

Besides evaluating the overall performance of MetaXT with relevant
baselines, we analyze the learned LTN for the cross-task learning for
sentiment analysis as follows:

For the sentiment analysis task transfer on Yelp and Amazon, as a-priori the target task (sentiment score is a scale of 1-5) is in
strongly linear correlation with the input source target task
(bi-polar sentiment), we seek to evaluate if the learned LTN is
actually able to transfer from a polarity sentiment correctly to the
multi-scale score. As such, we take the learned LTN and plot the pairs
of (input bi-polar sentiment, produced multi-scale score) as shown in
Figure \ref{fig:yelp_amazon_confusion_matrix} for Yelp and Amazon.


We can see that the learned LTN from MetaXT is able to figure out that
a source label ``thumbs-down'' is mostly likely to correlate with a
1-star rating in the target label space. (Note that the model itself
does not have access to such information when it's trained). There is
still some noise in the inferred target label distribution given a
source label, which could be explained by optimization difficulty
brought by the the bi-level optimization framework.
For an analysis of LTN behavior with the NER tasks please refer to Appendix B.

\subsubsection{Ablation studies}

Effectively, there are two novel ingredients in MetaXT: a) The model
architecture with a separate LTN besides a
standard multi-task model; b) The bi-level optimization strategy to
train both the main model and the LTN for best transferring the source
task examples to help the target task. We seek to
evaluate the effect the bi-level optimization strategy while keeping
the LTN, i.e., using the LTN as a third source of multi-task training
besides the source and target task data, by directly optimizing
$\min_{\alpha, \theta, v, w} \mathcal{L}_{train}$ where
$\mathcal{L}_{train}$ is training loss defined the same as Eq
(\ref{eq:loss_train}). Note that there is no meta-parameter in this
formulation as $\alpha$ is also part of the main parameters to
optimize. We term this model variant as XT due to its nature of
lacking the meta-learning formulation. At its core, XT is a multi-task
training baseline with the additional model capacities from the LTN. 

\begin{table}[t]
  \centering
  \caption{MetaXT vs XT - Yelp Dataset (Acc.  $\%$)}
  \begin{tabular}{cccccc}
    \toprule
    $k$& \thead{20} & \thead{50} &\thead{100} & \thead{200} & \thead{500} \\
    \midrule
    Multi-Task       & 21.8 & 31.1 & 37.7 & 43.8 & 47.7 \\\midrule    
    MetaXT          & \textbf{31.4} & \textbf{38.0} &\textbf{42.0} & \textbf{43.9} &\textbf{48.1}  \\
    XT          & 20.6 & 20.1 & 22.5 & 22.9 & 38.5  \\
    \bottomrule
  \end{tabular}
  \label{table:ablation_xt_yelp}
\end{table}

We evaluate XT against MetaXT on the Yelp and Amazon data sets on the
setting of transferring from bi-polar sentiment to multi-scale
ratings. Table \ref{table:ablation_xt_yelp} shows that without the
bi-level optimization formulation, XT's performance drops
significantly from MetaXT. Note that XT is worse than the Multi-Task
baseline, even with additional model parameters (from the LTN). This
is however expected as the pseudo-labels from the LTN is trusted as a
third source of information in training the main model, however
without the bi-level optimization, there's no way to ground the
generated labels from the LTN such that the low quality labels from
the LTN causes the model to dramatically decrease in
performance. Similar trends are observed on the Amazon data set (Table
\ref{table:ablation_xt_amazon}).

\begin{table}[t]
  \centering
  \caption{MetaXT vs XT - Amazon Dataset (Acc. $\%$)}
  \begin{tabular}{cccccc}
    \toprule
    $k$& \thead{20} & \thead{50} &\thead{100} & \thead{200} & \thead{500} \\
    \midrule
    Multi-Task       & 23.8 & 33.6 & 34.4 & 45.3 &49.9 \\\midrule    
    MetaXT          & \textbf{35.9} & \textbf{40.6} & \textbf{42.4} & \textbf{47.1}&\textbf{50.0} \\
    XT   & 20.5 & 19.8 & 26.4 & 23.5 &37.6 \\    
    \bottomrule
  \end{tabular}
  \label{table:ablation_xt_amazon}
\end{table}

\section{Related Work}
\label{sec:relatedwork}

\textbf{Transfer learning across tasks:} Adapting a pre-trained
language model to a specific NLP task is a  form of
transfer learning between the language modeling task (during pre-training) and the downstream task (during fine-tuning) that has become a defacto standard for many NLP tasks. However, it still leaves margin for improvement
by transferring between related downstream NLP tasks, especially for the few-shot setting where limited labeled data is available for a given task. Many NLP methods leverage multi-task learning to improve performance across related NLP tasks, e.g. ~\cite{ruder2019latent, balikas2017multitask, augenstein2018multi,pfeiffer2020mad} and loosely connected, such as transferring from POS tagging to NER or chunking~\cite{ruder2019latent}, transferring between POS tagging and neural machine translation~\cite{niehues2017exploiting}.

Multi-task learning methods typically use techniques like joint training ~\cite{augenstein2018multi} or splitting parameters into shared and private spaces ~\cite{liu-etal-2017-adversarial}. Most recently, task adaptors ~\cite{pfeiffer2020adapterfusion,pfeiffer2020mad}, small learnt bottleneck layers inserted within each layer of a pre-trained model, were successfully used for multi-task learning. We build on top of this line of work but we focus on cross-task transfer where limited labeled data is available for the target task. We show that our approach, using a bi-level optimization technique, can best leverage the data from an auxiliary task to optimize the performance on the target task.

\textbf{Meta-learning for task adaptation:} Another line of work
related to this paper is on meta-learning for multi-task
learning~\cite{finn2017model, chen2018meta}. MAML~\cite{finn2017model} aims to learn from a set of
training tasks and to generate well onto unseen testing tasks by
learning model initializations for faster
adaptation. MANN~\cite{santoro2016meta} builds on top of MAML by
explicitly leveraging a memory network to improve task level
generalization. Our setting is different in that first we assume
that the target task in observed in training data and second, we focus
on the few-shot aspect for the target task to explicitly highlight the
potential values of transferring from an auxiliary source task.


\textbf{Domain adaptation} Another closely related line of work is on
the domain adaptation and transfer, which looks to transfer and re-use
knowledge obtained from one domain to another, or to address the
intrinsic distribution shift underlying the
data~\cite{kouw2018introduction, teshima2020few,
  motiian2017few,hu2018mtnet, chidlovskii2016domain}. This line of
work typically focus on the transfer between different data spaces,
whereas in this paper, we mainly focus on the transfer between the
task spaces, specially for those with disparate label spaces. Though
we don't emphasize the potential differences of the underlying domains
between the source and target task, our method is applicable to such
settings, which is also validated by the experiments on the transfer
from POS tagging to MWE and from NER between two different German NER sets.
For a discussion on related work in the area of weak supervision please refer to Appendix C.

\section{Conclusions and Future Work}
\label{sec:conclusion}

In this paper, we study the problem of transfer learning across NLP
tasks from disparate label spaces in the few shot setting. Directly fine-tuning a pre-trained language model on the target task or
jointly training on both tasks tends to be sub-optimal due to limited labeled data. Instead, we devise a
label transfer network (LTN) to explicitly transform the available
source task labels into the target task label space such that the
target task model trained with these amended ``pseudo'' labels can
best perform on a separate portion of the target task labeled
set via a bi-level optimization framework termed MetaXT. We conduct experiments on four different
transfer settings across two different types of disparate label spaces
and empirical evaluations verify that MetaXT outperforms the baselines
or comparably to the best model, particularly on the few-shot
settings. A potential direction is to extend MetaXT onto
settings with multiple source tasks by building a universal LTN for all source tasks to transfer to the same target task space to leverage knowledge from multiple  sources simultaneously.

\section*{Ethical Considerations}
This work addresses cross task transfer between disparate label spaces. The proposed method is computationally intensive to some degree as it involves computation of second order gradients over pre-trained Language Models in order to compute the meta-gradients. This might impose a negative carbon footprint from training the described models. Future work on developing efficient meta-learning optimization methods and accelerating the meta-learning training procedures might help in alleviating this concern.

\bibliographystyle{acl_natbib}
\bibliography{main}

\newpage
\appendix

\section{Details of experimental setup}

For all methods, we use the same PLM as the
textual feature encoder for all data sets, i.e.,
BERT~\cite{devlin2018bert}. For the NER transfer dataset, the
pretrained German BERT models \footnote{\url{https://github.com/dbmdz/berts\#german-bert}} were used as the PLM. 
Hyper-parameters of all methods are
selected based on the model performance on a separate validation data
set. 
All models were trained with a batch size of 10.
For the Adapter Fusion models, we use the single-task adapters to train on the source and target tasks, and then adapt the model to the target task.
For the MetaXT
Model, the batch was split into two mini-batches for training and meta-training. 
To evaluate the performance of
the models in true low example target settings, the size of the
validation set was sized to be similar to the small training
set of the target task. For sequence classification tasks, the target training and the validation datasets were
balanced to have equal representation from all classes. The sequence tagging
dataset sizes were selected to have a minimum number of examples from
all classes. 
The sequence lengths for the granularity datasets were
capped at 128 tokens and the sequence lengths for the sequence tagging
datasets was capped at 64 (due to smaller average sentence sizes in the
overall dataset). 
We describe processing procedures specific to the
data sets in the following section. We implement all baseline methods
in PyTorch and code for MetaXT will be made publicly available.

\section{LTN Analysis - GermEval 2014}
\label{sec:appendix}
\begin{figure*}[t]
    \centering
    \includegraphics[width=0.95\textwidth]{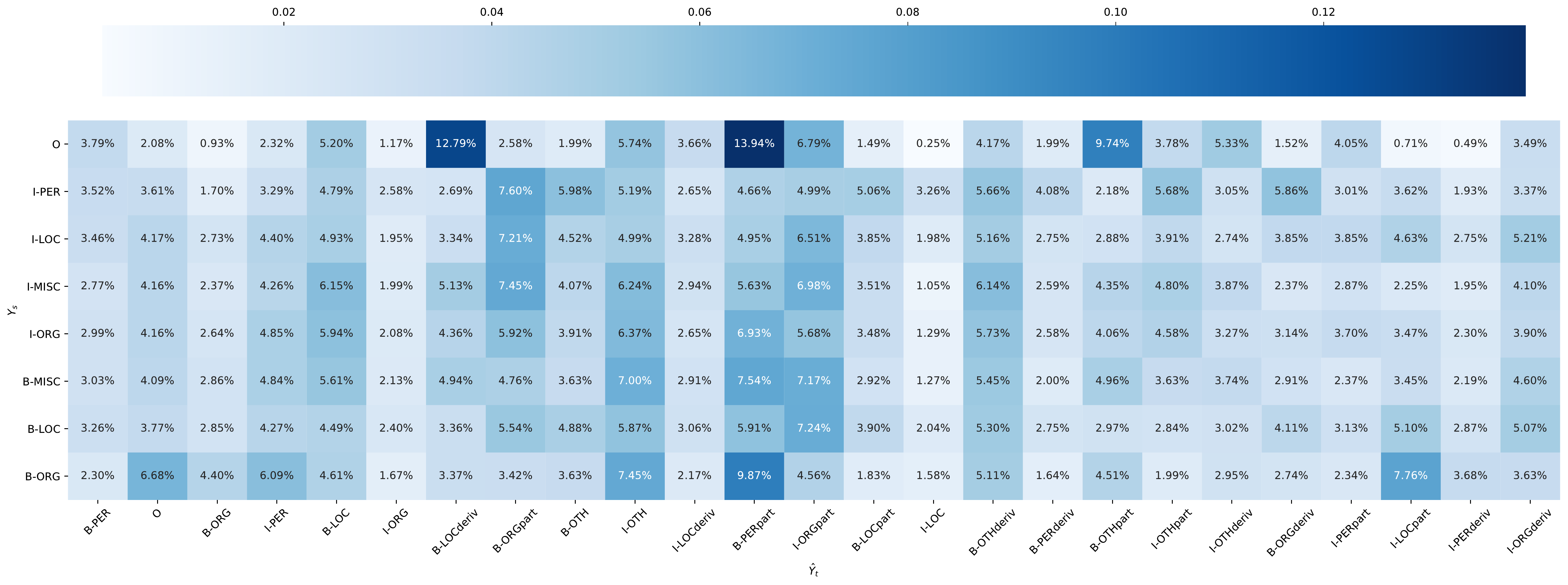}
    \caption{Distribution of \(\hat{Y}_t\) given \(Y_s\) for German NER tag transfer dataset run with 100 examples. Even with 50 target examples, the LTN is able to learn and reinforce some correlations with the target data. For example, the connection between B-ORGpart and Location and Misc information (I-LOC, I-MISC), and identification that label 'O' in the source label maps could map to many derived labels in the target NER label space.}
    \label{fig:ner_mltl_confusion_matrix}
\end{figure*}

The average distribution of learned LTN labels for transfer between different NER tag sets is shown in Figure \ref{fig:ner_mltl_confusion_matrix}. The tag sets for the target task are higher fidelity, and some of the derived labels correspond to the generic NER label 'O' of the source tag set. There are however, some tags that are common between the two (such as 'B-ORG', 'I-PER'). 
For some of the direct correspondence labels, the LTN is able to identify and reinforce the target label with additional examples. With others, such as 'I-PER', the LTN identifies a correspondence, but does not fully learn the direct correspondence. 
The LTN is able to identify that there are correlations between certain labels that are a consequence of annotation guidelines. For example, there is correlation between 'B-ORGpart' and 'B-LOC' due to the frequent specification of location details associated with organization text, and having location information commonly be part of organization names, such as "[Linke-Europaabgeordnete](ORGpart)"

\section{Related Work - Weak Supervision}
Given the problem scope in this paper, one might also attempt to treat the source task as weak
supervision signals for the target task and leverage existing work on
learning with weak supervision to address transfer learning on
tasks. However, one major hurdle that prevents applying such methods is that the existing work on weakly supervised learning is only applicable to the setting where the weak labels are from the same label space as the target task. Such work ranges from jointly learning with clean and weak labels~\cite{sukhbaatar2014training,tanaka2018joint,shu2020learning}, to learning to re-weight the weakly supervised labels~\cite{ren2018learning, shu2019meta, shu2020leveraging}, to aim to correct the weak or noisy labels as in ~\cite{patrini2017making,hendrycks2018using, zheng2021meta}. To transfer between disparate label spaces, the notion of weak supervision goes beyond existing work on weakly supervised learning, while the proposed MetaXT can also be viewed as an attempt to address this generalized notion of weak supervision defined from a source task.



\end{document}